\documentclass{article}

\usepackage{microtype}
\usepackage{graphicx}
\usepackage{subcaption}
\usepackage{booktabs} 

\usepackage{hyperref}

\usepackage[accepted]{icml2026}

\usepackage{amsmath}
\usepackage{amssymb}
\usepackage{mathtools}
\usepackage{amsthm}

\usepackage[capitalize,noabbrev]{cleveref}

\theoremstyle{plain}

\theoremstyle{definition}

\theoremstyle{remark}

\icmltitlerunning{Alignment-Aware Decoding}

\usepackage{amsmath,amsfonts,bm}

\def\eqref#1{equation~\ref{#1}}

\def\1{\bm{1}}

\DeclareMathAlphabet{\mathsfit}{\encodingdefault}{\sfdefault}{m}{sl}
\SetMathAlphabet{\mathsfit}{bold}{\encodingdefault}{\sfdefault}{bx}{n}

\newcommand{\E}{\mathbb{E}}

\newcommand{\KL}{D_{\mathrm{KL}}}

\DeclareMathOperator*{\argmax}{arg\,max}

\usepackage[utf8]{inputenc}
\usepackage{tabularx} 
\newcolumntype{Y}{>{\centering\arraybackslash}X}
\newcolumntype{C}[1]{>{\centering\arraybackslash}p{#1}}
\usepackage{multirow}
\usepackage{nicefrac}
\usepackage{bbding} 

\usepackage{pifont}

\usepackage{mdframed}
\usepackage{verbatim}
\usepackage{fvextra}

\newcommand{\sft}{\textsc{sft}}

\usepackage{pgfplots}
\pgfplotsset{compat=1.18}
\usepackage{url}
\usepackage{wrapfig}
\usepackage{float}
\usepackage{comment}

\usepackage{tikz}
\usetikzlibrary{arrows.meta,positioning,fit,calc}

\begin{document}

\twocolumn[
  \icmltitle{Alignment-Aware Decoding}

  \icmlsetsymbol{equal}{*}
  \begin{icmlauthorlist}
    \icmlauthor{Frédéric Berdoz}{eth}
    \icmlauthor{Luca A. Lanzendörfer}{eth}
    \icmlauthor{René Caky}{eth}
    \icmlauthor{Roger Wattenhofer}{eth}
  \end{icmlauthorlist}
  \icmlaffiliation{eth}{ETH Zurich, Switzerland}
  \icmlcorrespondingauthor{Frédéric Berdoz}{fberdoz@ethz.ch}

  \icmlkeywords{Machine Learning, ICML}

  \vskip 0.3in
]

\printAffiliationsAndNotice{Code available at \url{https://github.com/ETH-DISCO/alignment-aware-decoding}.}

\begin{abstract}
Alignment of large language models remains a central challenge in natural language processing. Preference optimization has emerged as a popular and effective method for improving alignment, typically through training-time or prompt-based interventions. In this paper, we introduce alignment-aware decoding (AAD), a method to enhance model alignment directly at inference. Theoretically, AAD can be interpreted as implicit reward optimization, yet it requires no specialized training beyond the standard DPO setup. Empirically, AAD consistently outperforms strong baselines across diverse alignment benchmarks and model scales. Moreover, in data-constrained settings, AAD can produce high-quality synthetic data to improve alignment under standard decoding, providing a practical solution when labeled data is limited.
\end{abstract}

\section{Introduction}

Large language models (LLMs) are the backbone of modern natural language processing, powering applications ranging from open-ended dialogue to complex reasoning tasks.  
Despite their impressive capabilities, aligning these models with human preferences remains a central challenge.  
Misaligned models can produce harmful, biased, or simply unhelpful outputs, motivating a growing body of work on alignment, i.e., the process of training models to better reflect human values and preferences \citep{ziegler2019fine,ouyang2022training,amodei2016concrete}.  
Alignment is typically performed during training, either through reinforcement learning from human feedback (RLHF) or more recent variants such as direct preference optimization (DPO)~\citep{rafailov2023direct}.  
While these methods can achieve strong empirical results, they tend to be sensitive to imperfect preference signals. In RLHF, this arises from errors in the learned reward model that can be exploited \citep{amodei2016concrete}, while in DPO it stems from noise in the preference data itself~\citep{rafailov2024scaling}.
To prevent over-optimization, the learned policy is typically constrained to remain close to a fixed reference model. This constraint ensures stability but also causes the optimal policy to inherit the biases of the reference model. This is because under this formulation, the learned policy is effectively trained as a reward model~\citep{rafailov2023direct}, and no longer as a policy that maximizes reward~\citep{rafailov2024from}.
An emerging alternative is \emph{inference-time alignment}, which steers model outputs at inference, without modifying parameters.  
Recent work explores emulated fine-tuning~\citep{mitchell2024an,liu2024tuning,xu2025genarm}, energy-based decoding~\citep{yuan2025inference,hong2025energybased}, and value-guided search~\citep{zhou2024weak,liu2024inference}, all of which leverage reward signals to bias generation.  
These methods offer flexibility when model weights are frozen or proprietary, but often require auxiliary models, complex search procedures, or carefully tuned hyperparameters to remain stable.

\begin{figure*}[t]
    \centering
    \includegraphics[trim=0 355 0 0, width=.98\linewidth]{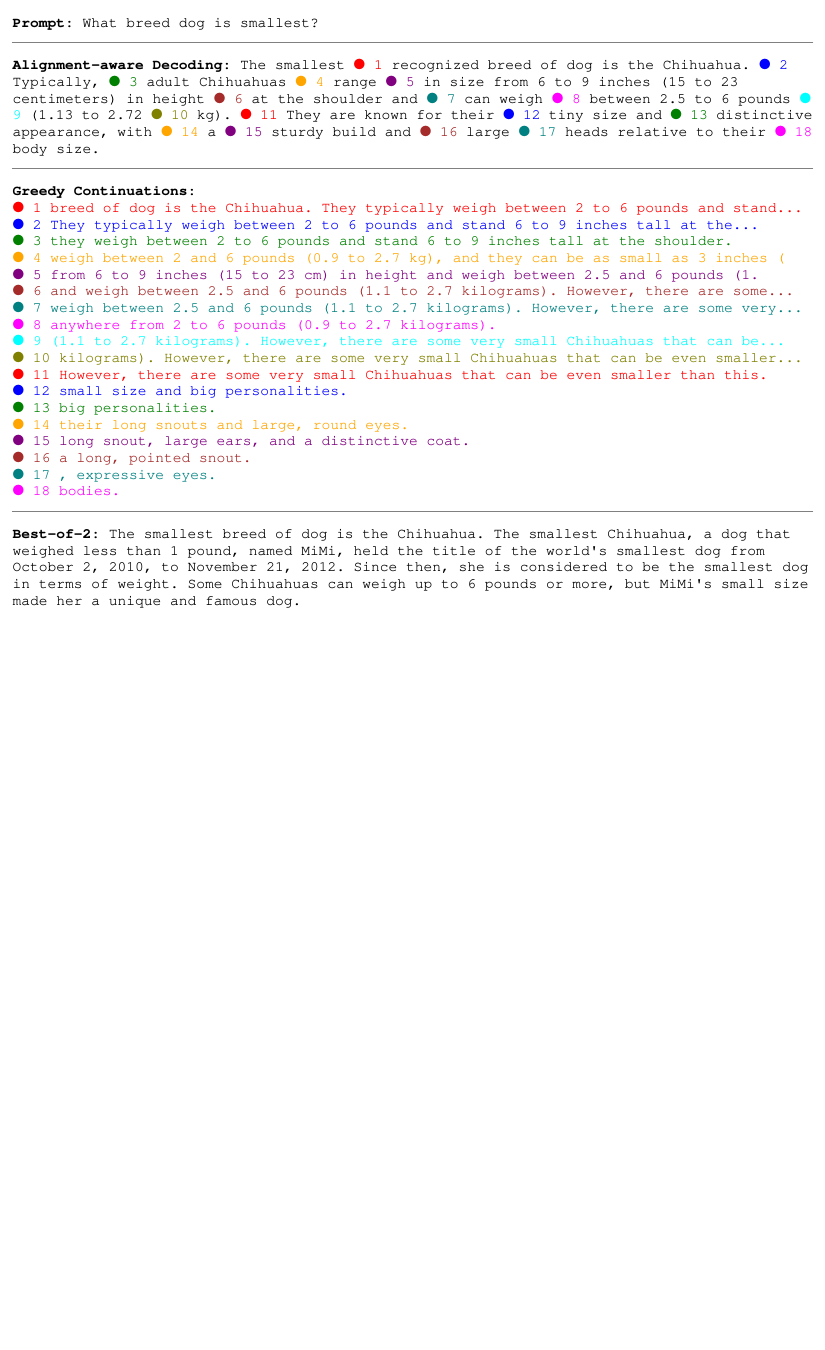}
    \caption{\textbf{Qualitative comparison of AAD against other decoding strategies}. Greedy continuations are generated by feeding the prompt together with the current AAD prefix back into the model and greedily selecting the next token, revealing where the greedy trajectory diverges from AAD. For instance, at \textcolor{magenta}{\textbullet\ 8}, given a context up to \texttt{[...] can weigh}, AAD generates \texttt{between 2.5 to 6 pounds [...]} while greedy generates \texttt{anywhere from 2 to 6 pounds [...]}. AAD identifies the Chihuahua as the smallest recognized breed of dog, making the distinction that it refers to an officially recognized classification, whereas the greedy continuation and Bo2 simply state \texttt{breed} without that nuance. AAD is also the only method that directly addresses size (the core of the prompt) by describing height and body proportions, while greedy and Bo2 focus mainly on weight. This highlights AAD’s advantage in preserving prompt adherence.}
    \label{dog}
\end{figure*}

In this paper, we introduce \emph{Alignment-Aware Decoding} (AAD), a simple method to reliably improve alignment directly at inference (see \cref{dog} for a qualitative example).
Our method leverages two distinct embedded features of the DPO-aligned model. First, its capacity to identify safe candidate tokens for the next decoding step via standard token likelihoods, and second, its ability to perform token-level credit assignment through the log-likelihood ratio with the reference model~\citep{rafailov2024from}. Intuitively, AAD exploits the alignment signal captured during preference optimization, which is often underutilized by standard decoding, and leverages the reference model at inference to mitigate biases it may have imparted to the aligned model, in a manner similar to methods that use a weaker (e.g., smaller) model to guide the decoding of a stronger model~\citep{li2023contrastive}.
Empirically, we demonstrate that AAD consistently improves alignment across diverse benchmarks and model scales under compute-equivalent conditions. Furthermore, when high-quality preference data or inference resources are scarce, AAD can generate high-value synthetic completions that can be fed back into the model through iterative DPO~\citep{pang2024iterative}, enabling stronger alignment without additional inference overhead.

We summarize our contributions as follows:
\begin{itemize}
    \item We introduce alignment-aware decoding (AAD), a simple inference-time method that uses the aligned model as a token reward function. Importantly, AAD requires no additional training, using only the reference model (before DPO) and the aligned model (after DPO).
    \item We demonstrate across multiple benchmarks and model scales that AAD consistently and significantly improves alignment over baselines under compute-equivalent conditions.
    \item We further demonstrate that AAD can be used to generate high-quality synthetic data to further improve the alignment of LLMs under standard decoding strategies.
\end{itemize}

\section{Related Work}

Recent efforts in aligning large language models (LLMs) with human preferences can be grouped into two broad categories: \emph{training-time alignment} and \emph{inference-time alignment}.  

\paragraph{Training-time alignment.}  
These approaches modify the model parameters to internalize the desired behavior directly during training. Reinforcement learning from human feedback (RLHF) is the standard paradigm for aligning LLMs~\citep{ziegler2019fine}, where a reward model is trained from human preferences and used to fine-tune the policy via a reinforcement learning algorithm such as proximal policy optimization (PPO)~\citep{schulman2017proximal}.
Direct preference optimization (DPO)~\citep{rafailov2023direct} and variants~\citep{hong2024orpo, azar2024general, ethayarajh2024kto, zhao2023slic} eliminate the reinforcement learning stage of RLHF by optimizing a simple objective that compares preferred and dispreferred outputs.  
Building on this idea, selective DPO~\citep{yang2024selective} improves sample efficiency by focusing the loss on key tokens with high preference signal.  
Weak-to-strong alignment~\citep{zhu2025weaktostrong} further extends the paradigm by using a smaller, already aligned reference model to guide the training of a larger base model, thereby transferring alignment without costly reward modeling.  
From a theoretical perspective, \citet{rafailov2024from} show that DPO training can be interpreted as learning a $Q$-function, which enables fine-grained credit assignment and provides a foundation for our method.

\paragraph{Inference-time alignment.}  
Instead of modifying the base model, these methods steer generation on the fly, offering flexibility when model weights are frozen or inaccessible.  
One prominent line of work is \emph{emulated fine-tuning} (EFT), where a reference-aligned model pair is used to define an implicit token-level reward function for decoding a third, unaligned base model~\citep{mitchell2024an}, effectively emulating its alignment at inference.
Such works include proxy alignment~\citep{liu2024tuning}, and GenARM~\citep{xu2025genarm}, which differ mainly in how the token reward signal is estimated.  
Energy-based decoding~\citep{yuan2025inference,hong2025energybased} takes a different angle by directly biasing generation toward low-energy regions of the reward and model’s logit landscape.  
Loosely related, \citet{liu2024decoding} introduce decoding-time realignment (DeRa), a decoding strategy that mimics DPO-trained models at different $\beta$ values without requiring retraining for each new value, and PAD~\citep{chen2025pad}, which integrates verbose preference signals into the reward.
Amulet~\citep{zhang2025amulet} is a training-free approach that formulates per-token decoding as an online learning problem to adapt generation to personalized preferences at test time.
Closer to our work are methods that employ explicit rewards along with lookahead search, such as DeAl~\citep{huang2024deal}, ARGS~\citep{khanov2024args}, controlled decoding~\citep{mudgalcontrolled}, and reward-guided beam search~\citep{deng2023reward}. By contrast, our method does not rely on a separate explicit reward function.
Tangential to our work are chunk-level value optimization methods that combine local search algorithm with external (implicit or explicit) value functions to select completions exhibiting the highest alignment. These include weak-to-strong decoding~\citep{zhou2024weak}, which generates candidate chunks with a base model and ranks them using an implicit value function derived from a reference-aligned model pair; IVG~\citep{liu2024inference}, which generates chunks via EFT and ranks them with a learned value function and PPO-guided Monte Carlo tree search~\citep{liu2024dont}, which reuses the value function obtained during PPO training to guide the search. 

\begin{table*}[t]

\centering
\small
\setlength{\tabcolsep}{4pt}
\renewcommand{\arraystretch}{0.85}
\caption{\textbf{Performance of AAD} across datasets, with decoding methods as rows and base models as columns.
Each cell reports the average oracle reward ($R$) and AAD's win rate ($W$) against the corresponding method.
Higher values indicate better alignment. AAD consistently achieves the highest rewards and win rate across all settings, demonstrating its strong alignment capability.}
\label{tab:results_aad_lora}
\begin{tabularx}{\textwidth}{p{2.0cm}*{4}{>{\centering\arraybackslash}X>{\centering\arraybackslash}X}}
\toprule
\textbf{Method} & \multicolumn{8}{c}{\textbf{Models} \& \emph{Datasets}} \\
\cmidrule(lr){2-9}
& \multicolumn{2}{c}{\textbf{Llama 3B}}
& \multicolumn{2}{c}{\textbf{Llama 8B}}
& \multicolumn{2}{c}{\textbf{Qwen 0.6B}}
& \multicolumn{2}{c}{\textbf{Qwen 4B}} \\
 & $R$ & $W$ & $R$ & $W$ & $R$ & $W$ & $R$ & $W$ \\
\midrule\midrule
& \multicolumn{8}{c}{{\emph{Ultrafeedback}}} \\
\cmidrule(lr){2-9}
Greedy SFT   & 0.58 & 0.86 & 0.87 & 0.85 & -0.88 & 0.80 & 0.22 & 0.80 \\
Greedy DPO   & 0.68 & 0.86 & 0.98 & 0.84 & -0.69 & 0.78 & 0.29 & 0.79 \\
Bo2          & 0.85 & 0.85 & 1.06 & 0.85 & -0.62 & 0.78 & 0.47 & 0.77 \\
EFT          & 1.04 & 0.83 & 1.27 & 0.81 & -0.19 & 0.67 & 0.58 & 0.73 \\
AAD (ours)   & \textbf{2.21} & - & \textbf{2.22} & - & \textbf{0.34} & - & \textbf{1.19} & - \\
\midrule
& \multicolumn{8}{c}{{\emph{Argilla}}} \\
\cmidrule(lr){2-9}
Greedy SFT   & 1.59 & 0.88 & 1.72 & 0.89 & -0.86 & 0.89 & 0.70 & 0.87 \\
Greedy DPO   & 2.48 & 0.86 & 2.55 & 0.87 & 0.12 & 0.80 & 1.37 & 0.82 \\
Bo2          & 3.02 & 0.84 & 3.16 & 0.86 & 0.68 & 0.77 & 1.94 & 0.78 \\
EFT          & 4.54 & 0.70 & 4.65 & 0.72 & 1.99 & 0.52 & 3.28 & 0.61 \\
AAD (ours)   & \textbf{5.64} &- & \textbf{5.90} & - & \textbf{2.33} & - & \textbf{3.84} & -\\
\midrule
&\multicolumn{8}{c}{{\emph{OpenRLHF Mixture}}} \\
\cmidrule(lr){2-9}
Greedy SFT   & 3.59 & 0.90 & 3.89 & 0.93 & 0.83 & 0.83 & 2.63 & 0.88 \\
Greedy DPO   & 4.54 & 0.88 & 4.93 & 0.89 & 1.74 & 0.76 & 3.56 & 0.79 \\
Bo2          & 5.34 & 0.83 & 5.60 & 0.85 & 2.42 & 0.68 & 4.48 & 0.69 \\
EFT          & 6.18 & 0.72 & 6.84 & 0.67 & 3.08 & 0.55 & 5.29 & 0.54 \\
AAD (ours)   & \textbf{7.28} & - &\textbf{7.60} & - & \textbf{3.42} & - & \textbf{5.45} & - \\
\midrule
& \multicolumn{8}{c}{{\emph{HHRLHF}}} \\
\cmidrule(lr){2-9}
Greedy SFT   & -1.89 & 0.62 & -1.13 & 0.61 & -1.36 & 0.65 & -0.53 & 0.64 \\
Greedy DPO   & -1.83 & 0.61 & -1.08 & 0.60 & -1.25 & 0.60 & -0.49 & 0.63 \\
Bo2          & -1.65 & 0.64 & -0.91 & 0.61 & -1.06 & 0.64 & -0.22 & 0.59 \\
EFT          & -1.74 & 0.61 & -0.98 & 0.57 & -1.12 & 0.57 & -0.47 & 0.64 \\
AAD (ours)   & \textbf{-0.97} & - & \textbf{-0.34} & - & \textbf{-0.61} & - & \textbf{-0.02} & - \\
\midrule
& \multicolumn{8}{c}{{\emph{Skywork}}} \\
\cmidrule(lr){2-9}
Greedy SFT   & 7.93 & 0.74 & 13.25 & 0.80 & -4.41 & 0.66 & 9.34 & 0.75 \\
Greedy DPO   & 8.45 & 0.72 & 13.64 & 0.78 & -3.73 & 0.66 & 9.54 & 0.74 \\
Bo2          & 9.04 & 0.74 & 14.15 & 0.76 & -5.18 & 0.73 & 9.35 & 0.76 \\
EFT          & 10.03 & 0.68 & 15.57 & 0.72 & -1.88 & 0.58 & 10.35 & 0.71 \\
AAD (ours)   & \textbf{13.71} & - & \textbf{19.27} & - & \textbf{-0.01} & - & \textbf{14.44} & - \\
\midrule
& \multicolumn{8}{c}{{\emph{Nectar}}} \\
\cmidrule(lr){2-9}
Greedy SFT   & 0.72 & 0.99 & 1.17 & 0.99 & -0.77 & 0.93 & 0.77 & 0.94 \\
Greedy DPO   & 1.45 & 0.98 & 2.12 & 0.99 & 0.09 & 0.84 & 1.45 & 0.85 \\
Bo2          & 2.15 & 0.95 & 2.64 & 0.93 & 1.07 & 0.70 & 1.99 & 0.74 \\
EFT          & 2.28 & 0.89 & 3.30 & 0.75 & 1.23 & 0.58 & 2.35 & 0.65 \\
AAD (ours)   & \textbf{3.63} & - & \textbf{3.70} & - & \textbf{1.68} & - & \textbf{2.71} & - \\
\bottomrule
\end{tabularx}
\end{table*}

\section{Background}
\label{sec:background}
\paragraph{Auto-regressive language modeling.} Let $\mathcal{V}$ denote the token vocabulary, and let $\pi$ denote an auto-regressive language model (LM) which, given a context $x$, generates a sequence $y$ with probability 
$\smash{\pi(y \mid x) = \prod_{t=1}^{|y|} \pi(y_t \mid x \circ y_{1:t-1}),}$ where $y_{1:t}$ denotes the prefix of $y$ up to and including position $t$, and $\circ$ denotes sequence concatenation ($y_{1:0}=\emptyset$ by convention).
Training $\pi$ typically involves three phases \citep{ziegler2019fine, ouyang2022training}: (i)~\emph{pretraining}, (ii) ~\emph{supervised fine-tuning} (SFT), and (iii)~\emph{preference optimization} (PO).  
During pretraining, the model is trained on large-scale unlabeled corpora to predict the next token given a prefix of text.
Then, this model is generally fine-tuned on curated, task-specific datasets through supervised learning, which improves its ability to follow instructions and generate useful outputs in more constrained settings (e.g., chatbot dialogue, summarization). For the remainder of this work, we denote by $\pi_{\textsc{sft}}$ the model obtained after SFT.
While such models can follow instructions, they often produce outputs that are suboptimal with respect to human values and preferences. PO further adapts $\pi_{\textsc{sft}}$ to better reflect these preferences.

\paragraph{Preference optimization.}
The goal of PO is to align the model with a conditional preference relation $\succ_x$, with $y_1 \succ_x y_2$ indicating that the completion $y_1$ is preferred over $y_2$ given the prompt $x$.
In practice, preference relations are typically modeled probabilistically using the Bradley-Terry (BT) model \citep{bradley1952rank}, which posits the existence of a scoring function $r^*$ that quantifies the quality of a prompt-completion pair $(x,y)$.
Specifically, with ${\sigma(z)=(1 + e^{-z})^{-1}}$ denoting the sigmoid function, the BT model defines the likelihood of $y_1$ being preferred over $y_2$ given $x$ as
\begin{equation}
\label{eq:BT}
p(y_1 \succ_x y_2) = \sigma\bigl(r^*(x, y_1) - r^*(x, y_2)\bigr),
\end{equation}
and therefore provides
a likelihood-based framework to train the LM on observed preferences. Starting from $\pi_{\textsc{sft}}$ and a prompt distribution $\rho$, the training objective of PO can be formulated as the KL-constrained optimization problem \citep{jaques2017sequence} of finding $\pi^*$ defined as:
\begin{equation}
    \label{eq:preference_optimization}
    \argmax_\pi \E_{x}\left[\E_{y}\left[r^*(x,y)\right] - \beta \KL\bigl( \pi_{|x}\Vert \pi_{\sft|x}\bigr)\right],
\end{equation}  
with $\beta>0$ a regularization hyperparameter and where the expectations are taken over $x\sim \rho$ and $y\sim \pi_{|x} := \pi(\cdot | x)$. The classical approach to solving \cref{eq:preference_optimization} is known as reinforcement learning from human feedback (RLHF), and proceeds in two steps \citep{ziegler2019fine}. First, a parametric reward model $r_\theta(x,y)$ is trained to minimize the negative log likelihood of observed preferences:
\begin{equation}
\label{eq:nllloss_reward}
\mathcal{L}(r_\theta ; \mathcal{D}) = - \frac{1}{|\mathcal{D}|} \sum_{ i=1}^{|\mathcal{D}|} \log \sigma\bigl(r_\theta(x^i, y_w^i) - r_\theta(x^i, y_l^i)\bigr) ,
\end{equation}
where $\mathcal{D}=\{(x^{i}, y_w^{i}, y_l^{i}) \mid x^{i} \sim \rho,\, y_w^{i} \succ_{x^{i}} y_l^{i}\}$ is a static preference dataset.
In a second stage, \cref{eq:preference_optimization} is approximately solved using policy gradient methods, such as PPO~\citep{schulman2017proximal}, on a parametric class of models. Despite their effectiveness, reinforcement learning algorithms are prone to reward hacking \citep{amodei2016concrete} and typically require generating many rollouts during training, which can be computationally expensive and unstable.
To address these challenges, \citet{rafailov2023direct} introduce \emph{direct preference optimization} (DPO) to directly approximate $\pi^*$ via a supervised objective. Formally, they note that the closed form solution of \cref{eq:preference_optimization} can be expressed in terms of the optimal policy as
\begin{equation}
\label{eq:optimal_policy}
\pi^*(y\mid x) = \frac{1}{Z(x)}\pi_\sft(y\mid x)\exp\left(\frac{1}{\beta}r^*(x,y)\right),
\end{equation}
with $\smash{Z(x; r^*)=\sum_{y'}\pi_\sft(y'\mid x)\exp(\frac{1}{\beta}r^*(x,y'))}$ the partition function. Rearranging the terms, they find that the $r^*$ must satisfy
\begin{equation}
\label{eq:reward_dpo}
r^*(x,y) = \beta \log \frac{\pi^*(y\mid x)}{\pi_\sft(y\mid x)} + \beta \log Z(x;r^*).
\end{equation}
The key idea of DPO is to eliminate the second stage of RLHF by directly minimizing \cref{eq:nllloss_reward} within a restricted reward class
$
\smash{r_\theta(x,y) = \beta \log \tfrac{\pi_\theta(y\mid x)}{\pi_\sft(y\mid x)}}.
$
 This choice ensures that $Z(x, r_\theta)=1$, such that, as per \cref{eq:reward_dpo}, $r_\theta(x,y) = r^*(x,y)$ if and only if ${\pi_\theta(y\mid x) = \pi^*(y\mid x)}$.  In other words, the model obtained after DPO, $\pi_\textsc{dpo}\approx \pi^*$, is simply a byproduct from training the reward model $r_\theta$ on the preference dataset $\mathcal{D}$.

\section{Method}

\paragraph{The aligned policy $\pi^*$ inherits the biases of $\pi_\textsc{sft}$.} The main motivation behind PO is that it increases the likelihood of completions with higher rewards, as shown in \cref{eq:optimal_policy}. However, counterintuitively, even the optimal analytical solution $\pi^*$ can sometimes favor a completion with a lower reward over one with a higher reward. To illustrate this, let $x$ be a prompt and $y_1, y_2$ any two completions satisfying $r^*(x,y_1) \geq r^*(x,y_2)$. From \cref{eq:optimal_policy}, we have
\begin{multline}
\log\frac{\pi^*(y_1\mid x)}{\pi^*(y_2\mid x)} =   \underbrace{\log \frac{\pi_\sft(y_1\mid x)}{\pi_\sft(y_2\mid x)}}_{:=\Delta_\textsc{sft}}  \\ + \frac{1}{\beta}\underbrace{\bigl(r^*(x,y_1) -r^*(x,y_2) \bigr)}_{:=\Delta_{r}}.\label{eq:proba_ratio}
\end{multline}
This implies that if $\smash{\Delta_\textsc{sft} < -\frac{1}{\beta}\Delta_r}$, then $\pi^*(y_1\mid x) \leq \pi^*(y_2\mid x)$ although $y_1$ is preferred over $y_2$ given $x$. In other words, the optimal model $\pi^*$ inherits the biases of \(\pi_\textsc{sft}\). Note that this is not due to reward hacking, as we only consider the exact reward $r^*$ in our derivation. This is consistent with the observation of \citet{rafailov2024from} that PO does not train a policy to directly maximize reward.

\paragraph{Token-level reward.} 
We propose to use $\pi_\textsc{dpo}$ together with $\pi_\textsc{sft}$ as an approximate token-level advantage function. Generating a completion $y$ given a context $x$ then amounts to maximizing
\[
r_\textsc{dpo}(x,y) = \beta \log \frac{\pi_\textsc{dpo}(y\mid x)}{\pi_\textsc{sft}(y\mid x)}.
\] 
Since exact maximization is intractable, we use greedy decoding with a token-level advantage function
\begin{equation}
\label{eq:token_reward}
A(v \mid x\circ y_{1:t}) = \log \frac{\pi_\textsc{dpo}(v\mid x\circ y_{1:t})}{\pi_\textsc{sft}(v\mid x\circ y_{1:t})} \qquad v \in \mathcal{V}.
\end{equation}

Decoding according to this ratio provides a more direct and human-aligned way to maximize reward than following $\pi_\textsc{dpo}$ directly. A detailed theoretical justification for this behavior can be found in \cref{app:aad_theory}. We omit $\beta$ as it does not change the ranking of candidate sequences.

\paragraph{Preventing over-optimization.}
Since DPO is trained on relatively small datasets, the advantage $A$ may be unreliable for low-probability tokens, and maximizing it without constraints at each decoding step typically produces degenerate completions. For instance, tokens that are essential for grammatical and semantic coherence might be assigned high probabilities by both $\pi_\textsc{dpo}$ and $\pi_\textsc{sft}$, making their ratio too small to be selected under the proposed decoding algorithm. Moreover, if $\pi_{\mathrm{sft}}$ assigns a small probability to a given token, even a tiny absolute increase from PO training can produce a large relative change, leading to spuriously high scores and numerical instabilities. To mitigate these issues, we take inspiration from contrastive decoding~\citep{li2023contrastive}, a decoding algorithm that uses a small model to boost the performance of a larger one, and similarly apply min-$\alpha$ filtering to the DPO probabilities $\pi_\textsc{dpo}$, restricting the alignment-aware decoding to plausible tokens only.

\paragraph{Proposed method: alignment-aware decoding (AAD).} Formally, alignment-aware decoding selects the token at position $t$ according to
\begin{equation}
\label{eq:aad}
y_{t} = \argmax_{v \in \mathcal{V}_\alpha(x \circ y_{1:t-1})} A(v|x \circ y_{1:t-1}),
\end{equation}
where $\mathcal{V}_\alpha(x \circ y_{1:t-1})\subseteq \mathcal{V}$ is the set of tokens satisfying
\begin{multline}
 \pi_\textsc{dpo}(v|x \circ y_{1:t-1}) \geq \alpha \max_{v'\in \mathcal{V}}\pi_\textsc{dpo}(v'|x \circ y_{1:t-1}),
 \label{eq:token_filtering}
\end{multline}
i.e., tokens over which alignment can safely be optimized. A pseudocode for AAD is presented in \cref{alg:aad}.

\begin{algorithm}[t]
\caption{Alignment-Aware Decoding (AAD)}
\label{alg:aad}
\begin{algorithmic}[1]
\REQUIRE DPO model $\pi_{\textsc{dpo}}$ with SFT version $\pi_{\textsc{sft}}$, prompt $x$, max length $T$, threshold $\alpha$
\STATE Initialize $y \leftarrow [\,]$
\FOR{$t = 1$ to $T$}
    \STATE Compute $\pi_{\textsc{dpo}}(\cdot \mid x \circ y_{1:t-1})$ and $\pi_{\textsc{sft}}(\cdot \mid x \circ y_{1:t-1})$
    \STATE Compute $A(\cdot \mid x \circ y_{1:t-1})$ 
    \STATE $p_{\max} \leftarrow \max_{v \in \mathcal{V}} \pi_{\textsc{dpo}}(v \mid x \circ y_{1:t-1})$
    \STATE $\mathcal{V}_\alpha \leftarrow \{ v \in \mathcal{V} \mid \pi_{\textsc{dpo}}(v \mid x \circ y_{1:t-1}) \ge \alpha \cdot p_{\max} \}$
    \STATE $y_t \leftarrow \arg\max_{v \in \mathcal{V}_\alpha} A(v \mid x \circ y_{1:t-1})$
    \IF{$y_t = \texttt{<eos>}$}
        \STATE\textbf{return } $y_{1:t-1}$
    \ENDIF
\ENDFOR
\STATE\textbf{return } $y_{1:T}$
\end{algorithmic}
\end{algorithm}

\begin{figure}[t]
    \centering
    \includegraphics[width=\linewidth]{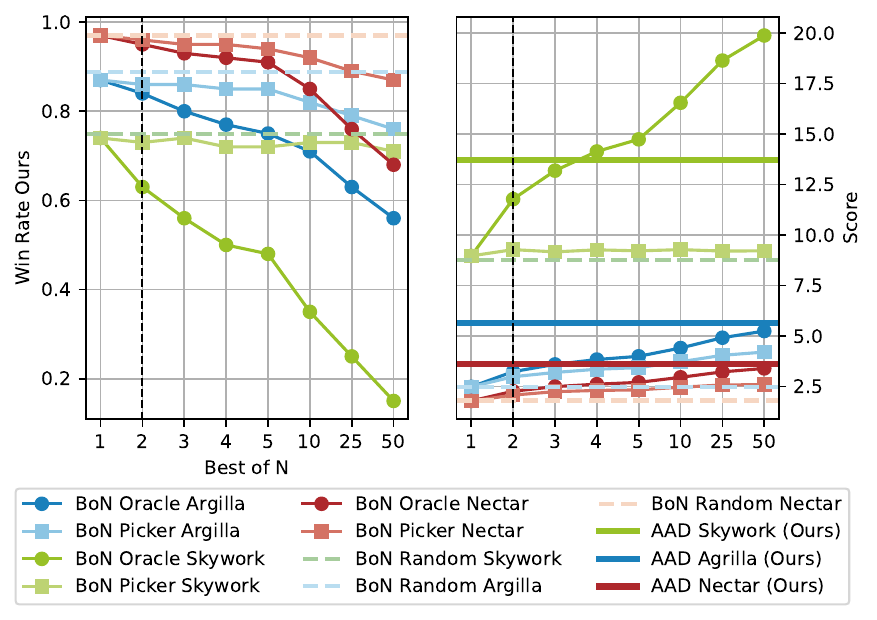}
    \caption{\textbf{AAD versus Bo$N$}. We evaluate AAD against three selection strategies on Argilla, Nectar and Skywork datasets for different values of $N$: (i) Bo$N$ using the oracle, (ii) Bo$N$ using the picker, and (iii) random selection among $N$ completions. AAD remains competitive even against Bo$N$-Oracle reward model, a setting that is by design unfavorable to AAD, since the oracle is used both for Bo$N$ selection and evaluation, whereas AAD only uses a model aligned on 10\% of the data. On Skywork, Bo$N$ reaches the performance of AAD for $N=4$ but requires roughly twice as much compute. On Argilla and Nectar, even $N=50$ fails to match AAD’s performance. The vertical dashed line indicates the point at which the computational cost of Bo$N$ matches that of our method. For the random selection baseline, we report only the mean performance across all test runs.}
    \label{bon}
\end{figure}

\section{Experimental setup}

\paragraph{Overview.} We conduct a series of experiments to evaluate the effectiveness of our method against several baselines. Each experiment begins with a preference dataset, which serves as the foundation for training both reward and aligned models. We split the data into a 90/10 training/evaluation set. An oracle reward model is trained on the full training split. In parallel, we subsample 10\% of the training split for two purposes: (i) training a picker reward model and (ii) aligning a SFT model $\pi_\textsc{sft}$ via DPO to obtain $\pi_\textsc{dpo}$. This setup allows us to simulate two conditions simultaneously: the availability of a strong oracle reward model for evaluation, and the scarcity of preference data, which is typically costly and difficult to obtain. The picker reward model is then used to select the highest-scoring continuation in methods such as best-of-$N$ (Bo$N$) sampling. For evaluation, we sample a fixed number of prompts from the validation split and generate continuations using both our method and the baselines. These continuations are scored with the oracle reward model. Evaluation metrics include (i) the win rate ($W$) of our method over a baseline, computed via pairwise continuation comparisons, and (ii) the average oracle reward ($R$) across all generated outputs. In addition, we also evaluate our method using the external AlpacaEval framework~\citep{alpaca_eval}. For reproducibility, we refer the reader to \cref{github}.

\paragraph{Datasets and reward models.}
For training and evaluation, we use preference datasets that are commonly adopted in reward modeling, including Ultrafeedback~\citep{ding2023enhancing}, Argilla~\citep{argilla}, the OpenRLHF Mixture~\citep{mixture,mixture2}, HHRLHF~\citep{hhrlhf}, Nectar~\citep{nectar}, and Skywork~\citep{skywork}. For the first 4 datasets, we train the reward models (pickers and oracles) using the training procedure detailed below.
We use Llama3.2~\citep{llama32} and Qwen3~\citep{qwen} base models and apply supervised fine-tuning (SFT) to obtain the SFT checkpoints used throughout the paper, following the workflow described by \citet{dong2024rlhf}.
For Skywork and Nectar, we do not train the oracles and instead follow a specialized evaluation protocol: prompts are drawn from the AlpacaEval dataset~\citep{alpaca_eval}, and scores are assigned using off-the-shelf oracle reward models trained externally on the respective datasets. Specifically, for Skywork we use the Skywork reward model (based on a Llama model~\citep{llama32}) and for Nectar we use the Starling reward model~\citep{nectar}. This ensures that the oracle has not been trained on the prompts used for evaluation. At the time of writing, the Skywork oracle is the top-performing model in Reward Bench~\citep{reward_bench}.

\begin{figure*}[t]
    \centering
    \begin{minipage}[t]{0.48\textwidth}%
        \centering
        \includegraphics[trim=12 -10 0 10, width=0.65\linewidth]{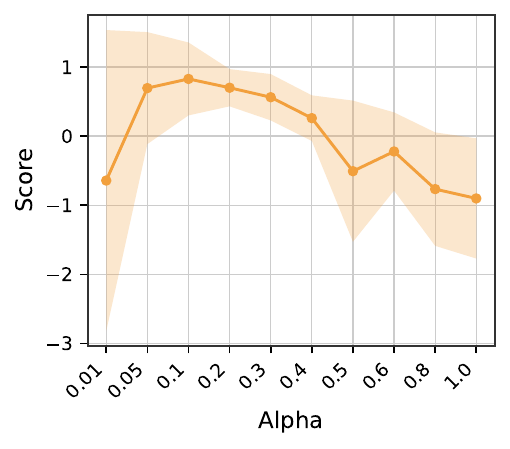}
        \caption{\textbf{Effect of the hyperparameter $\alpha$ in AAD}. Mean and standard deviation across the preference datasets Skywork, Nectar and Argilla, evaluated with a 3B LLaMA model fine-tuned with LoRA. Model performance exhibits a clear peak for $\alpha$ values in the range of approximately 0.1 to 0.2.}
        \label{fig:alpha}
    \end{minipage}%
    \hfill
    \begin{minipage}[t]{0.48\textwidth}%
        \centering
        \includegraphics[trim=0 5 0 0, width=0.8\linewidth]{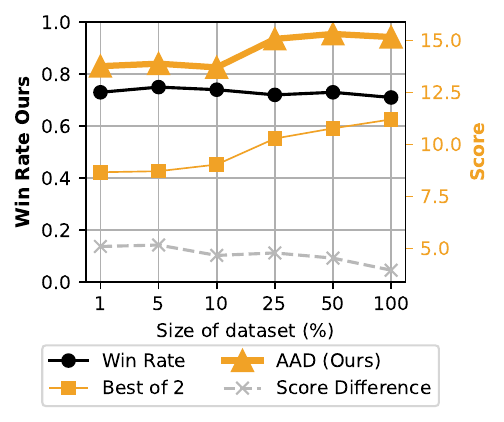}
        \caption{\textbf{Performance of AAD across different training dataset sizes} on the Skywork dataset. Results show that AAD consistently outperforms Bo2 at every data scale, providing clear evidence of its robustness in low-data regimes.}
        \label{perc}
    \end{minipage}
\end{figure*}

\begin{table}[t]
\centering
\small
\setlength{\tabcolsep}{2pt}
\renewcommand{\arraystretch}{1}
\caption{\textbf{AAD win rate on AlpacaEval} with default evaluator (GPT-4) across Skywork and Nectar \citep{alpaca_eval}. AAD consistently matches or outperforms baselines. \textbf{L} is short for Llama SFT and \textbf{Q} for Qwen SFT.}
\label{tab:alpacaeval}
\begin{tabularx}{0.99\linewidth}{l l *{4}{>{\centering\arraybackslash}X}}
\toprule
\textbf{Dataset} & \textbf{Method}
& \textbf{Llama 3B} & \textbf{Llama 8B} & \textbf{Qwen 0.6B} & \textbf{Qwen 4B} \\
\midrule
\multirow{4}{*}{\textit{Skywork}}
& Greedy SFT  & 0.77 & 0.79 & 0.74 & 0.76 \\
& Greedy DPO  & 0.76 & 0.77 & 0.73 & 0.75 \\
& Bo2         & 0.75 & 0.78 & 0.73 & 0.77 \\
& EFT         & 0.73 & 0.73 & 0.65 & 0.73 \\
\midrule
\multirow{4}{*}{\textit{Nectar}}
& Greedy SFT  & 0.80 & 0.82 & 0.52 & 0.61 \\
& Greedy DPO  & 0.76 & 0.76 & 0.44 & 0.54 \\
& Bo2         & 0.76 & 0.72 & 0.48 & 0.50 \\
& EFT         & 0.70 & 0.63 & 0.44 & 0.50 \\
\bottomrule
\end{tabularx}
\end{table} 
\paragraph{Training.}
We train both the pickers (for all datasets) and oracle reward models (except for Skywork and Nectar) using full fine-tuning with an additional classification layer, optimized under the Bradley-Terry loss detailed in \cref{eq:nllloss_reward}. Training is performed for two epochs. For the aligned models $\pi_\textsc{dpo}$, we also conduct two epochs of training, employing LoRA adapters \citep{lora}. Comprehensive training details are provided in \cref{train_details}. The accuracies of the oracle and picker reward models on the evaluation splits of the datasets are reported in \cref{acc_rew}.

\paragraph{Baselines.} For evaluation, we compare our method against four alternative decoding strategies that only use $\pi_\textsc{dpo}$, $\pi_\textsc{sft}$, or both: (i) greedy decoding with $\pi_\textsc{sft}$, (ii) greedy decoding with $\pi_\textsc{dpo}$, (iii)  Bo2 sampling with $\pi_\textsc{dpo}$, and (iv) a variation of EFT~\citep{mitchell2024an,liu2024tuning,rafailov2024from} using $\pi_\textsc{sft}$ for both the base and reference model, and setting $\beta = 4$, which has been found to perform the best across multiple settings.  For (iii), two candidate responses are generated with the aligned model via nucleus (top-$p$) sampling with $p = 0.9$~\citep{Holtzman2020The}, after which the picker reward model of the corresponding preference dataset selects the higher-scoring output. Both (iii) and (iv) entail a computational cost comparable to our method, whereas (i) and (ii) incur roughly half that cost.

\paragraph{Generation.}  
During AAD decoding, we set the token filtering parameter  $\alpha = 0.1$ as defined in \cref{eq:token_filtering} and ablated in \cref{fig:alpha}. This value of $\alpha$ is used unchanged across all 6 preference datasets and 4 model families reported in \cref{tab:results_aad_lora}, indicating that $\alpha = 0.1$ transfers across tasks and model scales without per-dataset tuning. Across all decoding methods, the \texttt{<user>} token is treated as an end-of-sequence marker.

\section{Results}

\paragraph{AAD consistently outperforms baselines.} The main results of our experiments are shown in \cref{tab:results_aad_lora}. Across both model families, AAD consistently outperforms the baselines by a substantial margin, achieving notably strong win rates with larger models. Remarkably, our method continues to deliver strong gains even when evaluated with external oracle reward models (Nectar and Skywork). On the AlpacaEval framework (see \cref{tab:alpacaeval}), our method also achieves mostly high win rates. We also test AAD against traditional contrastive decoding \citep{li2023contrastive}, where we use an amateur model instead of the SFT version in \cref{eq:proba_ratio}. The results in \cref{tab:lora-aad_vs_cd} strongly suggest that the SFT model is crucial for the alignment improvements achieved by AAD.
We provide additional results for AAD in \cref{sec:additional_results}. These include a detailed study of how AAD interacts with beam search, where we show that naive beam search often degrades alignment due to beam collapse, but can be stabilized via entropy thresholding and appropriate choice of $\alpha$. We further analyze the effect of beam width and demonstrate that, even under favorable settings, standard beam search provides limited gains compared to AAD. We also report results with fully fine-tuned models (beyond LoRA), confirming that the benefits of AAD persist under full fine-tuning, and study the effect of training duration, showing that AAD remains stable across epochs despite mild overtraining effects.

\begin{table}[h]
\centering
\small
\caption{
\textbf{Comparison between AAD and contrastive decoding (CD)} \citep{li2023contrastive}  on weaker reference models across the Argilla, Skywork, and Nectar datasets. All methods use LLaMA-8B DPO as the main model. CD subtracts logits from external LLaMA Instruct models (1B and 3B)~\citep{llama32}, treating them as generic ``amateur'' references, while AAD uses the 8B SFT model that the DPO model was trained from. Reported metrics include the reward-model score ($R$) and win rate ($W$) of AAD. AAD consistently achieves the highest reward across all datasets.}

\label{tab:lora-aad_vs_cd}
\begin{tabularx}{\linewidth}{l *{6}{>{\centering\arraybackslash}X}}
\toprule
\textbf{Method} & \multicolumn{2}{c}{\emph{Argilla}} & \multicolumn{2}{c}{\emph{Skywork}} & \multicolumn{2}{c}{\emph{Nectar}} \\
\cmidrule(lr){2-3} \cmidrule(lr){4-5} \cmidrule(lr){6-7}
 & $R$ & $W$ & $R$ & $W$ & $R$ & $W$ \\
\midrule
CD 1B Instruct & 2.79 & 0.85 & 13.08 & 0.81 & 2.65 & 0.90 \\
CD 3B Instruct & 2.46 & 0.86 & 10.88 & 0.88 & 2.51 & 0.93 \\
AAD (ours) & \textbf{5.9} & - & \textbf{19.27} & - & \textbf{3.7} & - \\
\bottomrule
\end{tabularx}
\end{table}

\paragraph{AAD with open-source DPO models.} We evaluate AAD on OLMo-2-7B-DPO and OLMo-2-13B-DPO~\citep{olmo2}, two open-source LLMs trained with full-parameter fine-tuning on a large preference mixture. For each model, we use the matching OLMo-2 SFT checkpoint as $\pi_\textsc{sft}$ and the OLMo-2 reward model for evaluation. Under reward-model evaluation (\cref{tab:olmo2}), AAD outperforms all four baselines at both 7B and 13B scales, with a win rate of 0.98 against the SFT baseline. AAD beats greedy DPO in 71.5\% of pairwise comparisons at 7B and 75.2\% at 13B, indicating that the gains hold under a reward-model-free evaluator.

\begin{table}[h]
\centering
\small
\caption{\textbf{AAD with open-source DPO models.} Performance of AAD on OLMo-2-7B-DPO and OLMo-2-13B-DPO~\citep{olmo2}, using the matching OLMo-2 SFT checkpoint as $\pi_\textsc{sft}$ and the OLMo-2 reward model. Columns $R$ and $W$ are the OLMo-2 reward score and the AAD win rate against the SFT baseline. Column AE is the AAD AlpacaEval win rate against the corresponding baseline under a GPT-4-turbo judge over 805 prompts.}
\label{tab:olmo2}
\setlength{\tabcolsep}{3pt}
\begin{tabularx}{\linewidth}{l *{6}{>{\centering\arraybackslash}X}}
\toprule
& \multicolumn{3}{c}{\textbf{OLMo-2-7B}} & \multicolumn{3}{c}{\textbf{OLMo-2-13B}} \\
\cmidrule(lr){2-4} \cmidrule(lr){5-7}
\textbf{Method} & $R$ & $W$ & AE & $R$ & $W$ & AE \\
\midrule
Greedy SFT & 1.51 & --   & 91.6 & 5.02 & --   & 93.9 \\
Greedy DPO & 3.27 & 0.94 & 71.5 & 6.87 & 0.95 & 75.2 \\
Bo2        & 3.44 & 0.97 & 64.6 & 6.94 & 0.96 & 65.2 \\
EFT        & 3.50 & 0.96 & 58.1 & 6.97 & 0.96 & 55.8 \\
AAD (ours) & \textbf{3.84} & 0.98 & -- & \textbf{7.22} & 0.98 & -- \\
\bottomrule
\end{tabularx}
\end{table}

\paragraph{Human evaluation.} To assess whether AAD's gains hold beyond reward-model evaluation, we conducted a head-to-head human study on responses from a 3B LLaMA model trained on the Skywork dataset. Results in \cref{tab:human_eval} show AAD wins between 58.7\% (versus greedy DPO) and 73.5\% (versus Bo2) of pairwise comparisons. Under Bradley-Terry Elo, AAD reaches 1610.1, ranking first by a margin of 85.7 points over the next baseline.

\begin{table}[h]
\centering
\small
\caption{\textbf{Human evaluation of AAD against baselines.} 502 pairwise judgments across 40 human raters on responses sampled from a 3B LLaMA model trained on the Skywork dataset. AAD win rate is the percentage of judgments where raters preferred AAD over the baseline. Elo is computed from pairwise outcomes via the Bradley-Terry model.}
\label{tab:human_eval}
\begin{tabularx}{\linewidth}{p{2cm}YY}
\toprule
\textbf{Method} & \textbf{AAD win rate} & \textbf{Elo} \\
\midrule
AAD (ours) & --     & \textbf{1610.1} \\
Greedy DPO & 58.7\% & 1524.4 \\
EFT        & 70.8\% & 1513.2 \\
Greedy SFT & 71.7\% & 1436.2 \\
Bo2        & 73.5\% & 1416.2 \\
\bottomrule
\end{tabularx}
\end{table}

\paragraph{Correspondence between Bo$N$ and AAD.}
In Bo$N$ sampling, the expected reward of the selected sequence increases as $N$ grows, since sampling more candidates raises the likelihood of obtaining a higher-scoring response by chance. \cref{bon} shows that our method remains competitive even when compared against Bo$N$ sampling with the oracle reward model, despite the oracle being trained on ten times more data than $\pi_\textsc{dpo}$, and despite the oracle being also used for the evaluation.

\begin{figure*}[t]
    \centering
    \begin{minipage}[t]{0.49\textwidth}
        \centering
        \includegraphics[width=0.74\linewidth]{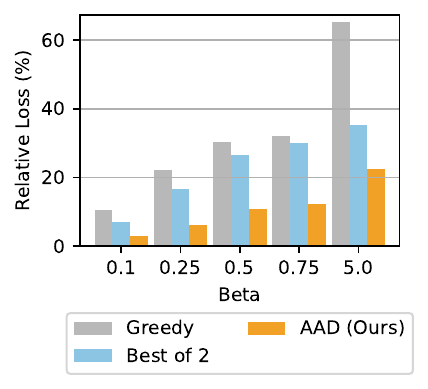}
        \caption{\textbf{Relative alignment loss} of the oracle score $R$ on the Argilla dataset as a function of the DPO regularization parameter $\beta$, with baseline performance established at $\beta = 0.05$. As expected, across all strategies, larger $\beta$ values reduce alignment, but AAD consistently shows the lowest relative loss, demonstrating greater hyperparameter robustness compared to baselines. This behavior stems from the fact that $r^*$ is $\beta$-independent, but $\pi^*$ is not, as seen in \cref{sec:background}.}
        \label{fig:b_rel}
    \end{minipage}%
    \hfill
    \begin{minipage}[t]{0.49\textwidth}
        \centering
        \includegraphics[width=0.74\linewidth]{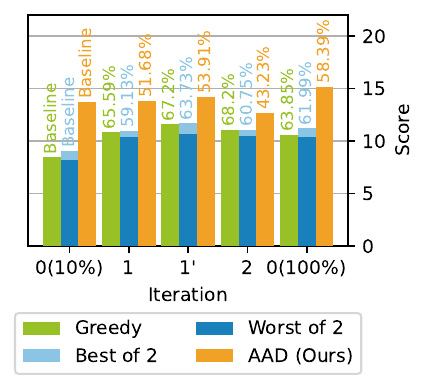}
        \caption{\textbf{Effect of iterative DPO}. Results show that iterative DPO using AAD-generated data substantially improves alignment, approaching full-dataset performance (100\%) with only 10\% of the original data. Win rates against the original $\pi_\textsc{dpo}$ (baseline), using the same decoding scheme, are shown above the bars. Iteration $i$ indicates the average oracle score of a model that has undergone DPO using AAD-generated data, initiated from $\pi_\textsc{sft}$ for $i=1,2$ or $\pi_\textsc{dpo}$ for $i=1'$.}
        \label{iterative}
    \end{minipage}
\end{figure*}

\paragraph{AAD performs strongly under data scarcity.} To assess our method in different data regimes, we train a series of picker reward models and aligned models on the Skywork dataset, gradually increasing the training data up to the full training split. We then evaluate our method against Bo2 sampling using the oracle. Results are shown in \cref{perc}. Note that the 100 \% mark in our plots does not represent the entire dataset used to train the external Skywork reward model, as we only trained on the 90\% training split and kept 10\% for evaluation. Interestingly, AAD’s win rate remains relatively consistent, suggesting that its performance generalizes across different data regimes.

\paragraph{Effect of DPO regularization parameter $\beta$.}
\label{iter}
The $\beta$ parameter constitutes a critical regularization hyperparameter in DPO training. To assess its influence on our method, we establish baseline performance at $\beta$ = 0.05 and evaluate the relative loss of models trained with $\beta$ values of 0.1, 0.25, 0.5, 0.75, and 5.0. We conduct these experiments on the Argilla dataset. The evaluation is conducted under three decoding strategies: Bo2 sampling, greedy decoding and our method. The corresponding results are presented in \cref{fig:b_rel}. Across all strategies, larger $\beta$ values are associated with reduced alignment performance. Nevertheless, our decoding method consistently exhibits the lowest relative loss, indicating greater robustness and stability compared to the alternative approaches.

\paragraph{Overcoming data scarcity with iterative DPO.} Since AAD appears to generate data with high alignment, we investigate if this data can be used to further train the aligned model. To this end, we implement a version of iterative DPO~\citep{pang2024iterative}. We begin with our model $\pi_\textsc{dpo}$, trained solely on 10\% of the original preference dataset (0th iteration), and using a Llama SFT model for $\pi_\textsc{sft}$. In the first iteration, we construct a synthetic preference dataset using the prompts of the subsampled dataset, and by pairing completions as follows: chosen samples are generated with AAD, while rejected samples are produced via nucleus sampling on $\pi_\textsc{dpo}$ with hyperparameter 0.9. We then retrain DPO alignment on this synthetic dataset in two variants: (i) starting from the base model (1st iteration) and (ii) starting from the model already aligned on the 10\% preference dataset (1' iteration). We further extend this process with a second iteration. Here, we retain the rejected samples from the previous step and generate new chosen samples using our method in combination with the DPO model trained during the 1st iteration. This produces a new synthetic dataset, which is again used to retrain DPO alignment from the base model (2nd iteration). Results shown in \cref{iterative} highlight the significant benefits of iterative DPO. Remarkably, even with only 10\% of the preference data, this method nearly closes the gap with a model trained on the full dataset.

\section{Conclusion}

We introduce \emph{alignment-aware decoding} (AAD), a decoding strategy that treats a DPO-trained model as a token-level reward function. AAD performs on-the-fly implicit reward optimization without additional training or external models. Across multiple datasets and model families, we show that AAD consistently improves alignment while maintaining efficiency comparable to standard decoding. AAD can also generate high-quality synthetic aligned data, enabling iterative preference optimization under data scarcity. While AAD improves alignment, there are limitations; it requires two forward passes per token, as well as access to the original SFT model. Our main claim is restricted to the standard SFT+DPO pipeline. As reported in \cref{tab:ppo}, combining AAD with PPO-trained models yields mixed results because PPO directly optimizes the policy against an external reward model and does not preserve the SFT-DPO probability gap that AAD exploits. Future directions include combining AAD with more sophisticated search strategies, exploring adaptive token filtering and entropy-based thresholds, and extending to other modalities such as image generation. Overall, we hope this work motivates further research on inference-time alignment methods that are both theoretically grounded and practically deployable.

\section*{Impact Statement}
This paper presents work whose goal is to advance the field of machine learning by improving the alignment of large language models at inference time. Improved alignment may reduce harmful or unhelpful model behavior in downstream applications, and may help practitioners deploy models more safely when training-time interventions are limited. At the same time, any method that alters model behavior can be misused, for example to enforce biased preferences, enable manipulation, or create misleading outputs that appear more persuasive. We therefore encourage careful evaluation, transparency about intended use, and deployment practices that include human oversight and monitoring.

\bibliographystyle{icml2026}
\bibliography{references}

\newpage
\onecolumn
\appendix
\section{Appendix}

\subsection{Detailed theoretical justification of AAD}
\label{app:aad_theory}
Our theoretical justification is based on the approach of \citet{rafailov2024from}. First, we model sequence generation as a token-level Markov Decision Process (MDP) $\mathcal{M}=(\mathcal{S}, \mathcal{V}, f, r_H, \rho)$, where $\mathcal{S}$ is the set of partial sequences $s_t = x \circ y_{1:t}$, $\mathcal{V}$ is the vocabulary of tokens, and the dynamics ${f(s,a) = s \circ a}$ deterministically append the selected token (action) $a$ to the current sequence $s$. The initial state distribution $\rho$ corresponds to prompts $x$, and the reward $r(s_t, a_t)$ defines the optimization problem. By convention, it is zero for all action if $s_t$ contains an end-of-sequence token \texttt{<eos>}. Typically, $r_H$ is such that $\sum_{t=0}^T r_H(s_t,a_t)=r^*(s_0,a_{0:T})$ reflects some sort of human preference for the complete sequence according to \cref{eq:BT}, with $T$ the index of the first \texttt{<eos>} token. Within this framework, one can rewrite the PO objective in \cref{eq:preference_optimization} as

\begin{align}
\label{eq:preference_optimization_token}
\pi^*&=\argmax_\pi \E_{x\sim \rho}\left[\E_{y\sim \pi(\cdot | x)}\left[r^*(x,y)\right] - \beta \KL\bigl( \pi(\cdot \mid x) \Vert \pi_\sft(\cdot \mid x)\bigr)\right]\\
& = \arg\max_\pi \mathbb{E}_{x \sim \rho,\, y \sim \pi(\cdot \mid x)} \Big[ r^*(x,y) - \beta \big( \log \pi(y \mid x) - \log \pi_\sft(y \mid x) \big) \Big] \\
& = \arg\max_\pi \mathbb{E}_{s_0 \sim \rho,\, \tau \sim \pi(\cdot \mid s_0)} \left[ \sum_{t=0}^T (r_H(s_t,a_t) + \beta \log \pi_\sft(a_t \mid s_t))\right] - \beta \mathbb{E}_{x \sim \rho} \mathcal{H}\big(\pi(\cdot \mid x)\big) \\
& = \arg\max_\pi \mathbb{E}_{s_0 \sim \rho,\, \tau \sim \pi(\cdot \mid s_0)} \left[ \sum_{t=0}^T r_\textsc{kl}(s_t,a_t)\right] - \beta \mathbb{E}_{x \sim \rho} \mathcal{H}\big(\pi(\cdot \mid x)\big),
\end{align} 
where $r_\textsc{kl}(s_t,a_t) = r_H(s_t,a_t) + \beta \log \pi_\sft(a_t \mid s_t)$ is the effective reward of the constrained MDP. We use the notation $\tau, s, a$ and $x,y$ for token-/sequence-level quantities, respectively.
Following \citet{ziebart2010modeling, rafailov2024from}, the fixed point solution of \cref{eq:preference_optimization_token} is given by 
\begin{equation}
\label{eq:fixed_point_solution}
\pi^*(a_t\mid s_t) = \exp\left(\frac{Q_\textsc{kl}^*(s_t, a_t) - V_\textsc{kl}^*(s_t)}{\beta}\right)
\end{equation}
with
\begin{align}
Q_\textsc{kl}^*(s_t, a_t) & = r_\textsc{kl}(s_t, a_t) + V_\textsc{kl}^*(s_{t+1}), \label{eq:Q_kl}\\
V_\textsc{kl}^*(s_t) & = \beta \log \sum_{a \in \mathcal{V}} \exp\left(\frac{Q_\textsc{kl}^*(s_t, a)}{\beta}\right).
\end{align}
Substituting \cref{eq:Q_kl} into \cref{eq:fixed_point_solution}, we get
\begin{equation}
\label{eq:dpo}
\pi^*(a_t\mid s_t) = \exp\left(\frac{r_\textsc{kl}(s_t, a_t) + V_\textsc{kl}^*(s_{t+1}) - V_\textsc{kl}^*(s_t)}{\beta}\right)
\end{equation}
Taking the logarithm of both sides and applying the definition of $r_\textsc{kl}$, we can rearrange to obtain
\begin{equation}
\label{eq:proba_ratio_derived}
\log \frac{\pi^*(a_t\mid s_t)}{\pi_\textsc{sft}(a_t\mid s_t)} = \frac{r_H(s_t, a_t) + V_\textsc{kl}^*(s_{t+1}) - V_\textsc{kl}^*(s_t)}{\beta},
\end{equation}
which motivates the token-level score in \cref{eq:token_reward}. As highlighted by \cref{eq:dpo} and \cref{eq:proba_ratio_derived}, decoding according to this ratio provides a more direct way to maximize the human-aligned reward $r_H$ compared to decoding with $\pi^*$. However, since DPO is typically trained on small datasets, the log ratio $\log \frac{\pi_\textsc{dpo}(a_t\mid s_t)}{\pi_\textsc{sft}(a_t\mid s_t)}$ might not be properly fitted for low-probability tokens unseen during training. To prevent selecting such tokens, we only consider tokens that have high probability under $\pi_\textsc{dpo}$ (as per \cref{eq:token_filtering}), preserving fluency while still favoring tokens (actions) with high human-aligned reward $r_H$.

\subsection{Accuracies of reward models}
\label{acc_rew}
In \cref{tab:reward_accuracy}, we report the accuracies of the picker and oracle reward models on the evaluation sets across all datasets.
\begin{table}[H]
\centering
\small
    \caption{\textbf{Accuracy of the reward models} trained on the different preference datasets. Oracles are trained on the full training split, and pickers on a 10\% subset.}
\label{tab:reward_accuracy}
\begin{tabularx}{\linewidth}{>{\centering\arraybackslash}p{6cm}>{\centering\arraybackslash}X>{\centering\arraybackslash}X}
\toprule
 \textbf{Dataset}   & \textbf{Accuracy Oracle (\%)} & \textbf{Accuracy Picker (\%)} \\
\midrule
Ultrafeedback \citep{ding2023enhancing}       & 76.2 &  69.5 \\

Argilla \citep{argilla}        & 92.3 &  82.5 \\

OpenRLHF Mixture \citep{mixture2}  & 85.6 &  77.9 \\

HHRLHF \citep{hhrlhf}          & 70.2 &  62.4 \\

Nectar \citep{nectar}      & external &  93.0 \\

Skywork  \citep{skywork}         & external &  78.1 \\
\bottomrule
\end{tabularx}

\end{table}

\subsection{Training details}
\label{train_details}
In this section we provide the training configurations and implementation details for the models used in our experiments.
\paragraph{Reward models.}  
Both the oracle reward models and the picker reward models are trained under identical hyperparameter settings:
\begin{itemize}
  \item Optimizer: AdamW 
  \item Batch size: 64
  \item Learning rate: $5 \times 10^{-6}$
  \item Training epochs: 2
  \item Gradient clipping: 1.0
  \item Precision: mixed-precision (bfloat16)
\end{itemize}

\paragraph{Aligned Model (DPO).}  
The aligned model $\pi_\textsc{dpo}$ is obtained by fine-tuning the base SFT model $\pi_\textsc{sft}$ using DPO on the 10\% subset. The training configuration is as follows:
\begin{itemize}
  \item Optimizer: AdamW with linear decay and linear warmup
  \item Batch size: 32
  \item Learning rate: $1 \times 10^{-6}$
  \item Warmup ratio: 0.1
  \item Weight decay: 0.1
  \item Training epochs: 2
  \item Gradient clipping: 1.0
  \item DPO coefficient ($\beta$): 0.1 (except for the experiment shown in \cref{fig:b_rel})
  \item Precision: mixed-precision (bfloat16)
\end{itemize}

\paragraph{LoRA Configuration.}  
To enable parameter-efficient fine-tuning, LoRA adapters are integrated into the DPO training pipeline with the following settings:
\begin{itemize}
  \item Rank ($r$): 64
  \item Alpha: 128
  \item Dropout: 0.05
  \item Target modules: attention projections (query, key, value)
\end{itemize}

\newpage
\subsection{Additional results}
\label{sec:additional_results}

\paragraph{Stabilizing beam search via entropy thresholding.}

\begin{wrapfigure}{r}{0.5\textwidth}
    \centering
    \vspace{-1em}
    \includegraphics[width=0.9\linewidth]{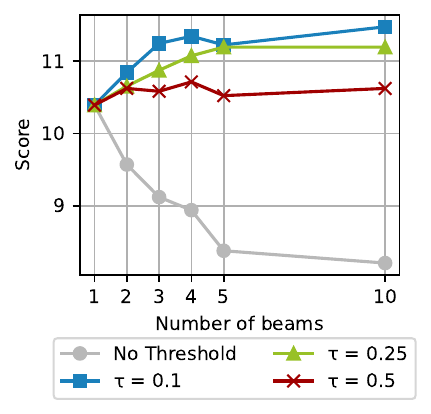}
    \caption{\textbf{Effect of beam size and entropy threshold} on performance for the Skywork dataset with $\alpha = 0.7$. Without entropy thresholding, scores rapidly degrade as the number of beams increases due to beam collapse. This mechanism enables larger beam sizes to yield improved alignment, while also reducing the computational cost compared to standard beam search.}
    \label{fig:beam}
    \vspace{-1em}
\end{wrapfigure}

We also investigate if we can use beam search on the token reward defined in \cref{eq:token_reward}, rather than simply greedy maximization. Beam search typically suffers from beam collapse, and increasing the number of beams does not always improve generations, a phenomenon reminiscent of inference-time over-optimization. However, we find that (i) increasing $\alpha$ and (ii) introducing an entropy threshold can make beam search beneficial in some cases. The key observation is that certain tokens are highly predictable and thus are assigned high probability by both $\pi_\textsc{sft}$ and $\pi_\textsc{dpo}$. In such cases, applying our score difference may incorrectly override an obvious continuation. To prevent this, we only apply our scoring adjustment when the aligned model is uncertain, that is, when the predictive entropy exceeds the threshold $\tau$. In practice, this is equivalent to setting $\pi_\textsc{sft}(y'\mid x)= \nicefrac{1}{|\mathcal{V}_\alpha(x)|}$ for every token $y'\in \mathcal{V}_\alpha(x)$ when the entropy  $\sum_{y' \in \mathcal{V}_\alpha(x)} -\pi_\textsc{sft}(y' \mid x) \log \pi_\textsc{sft}(y' \mid x) \leq \tau$. The results in \cref{fig:beam}, obtained on the Skywork dataset with $\alpha = 0.7$, show that without entropy thresholding, scores rapidly degrade as the number of beams increases. By contrast, introducing the threshold stabilizes performance and makes beam search beneficial.

\paragraph{Fully fine-tuned results.} In \cref{tab:results_aad_full}, we present an extension of our main results table with two aligned models, trained and evaluated under the same procedure as in \cref{tab:results_aad_lora}, with the exception that we perform full fine-tuning instead of using LoRA adapters. The results yield similar conclusions about the effectiveness of AAD. One model is trained from a 1B Llama SFT model and the other from a 3B Llama SFT model~\citep{llama32}. The gap between AAD and the baselines is narrower than in the LoRA setting. This is expected because full fine-tuning lets the DPO model absorb more of the preference signal into its parameters, leaving less residual alignment signal for AAD to exploit at inference. This matches the theoretical framing in \cref{sec:background}, where AAD's gain grows when the SFT prior biases the aligned policy away from the reward-optimal output.

\begin{figure}[ht]
    \centering
    \includegraphics[width=0.7\linewidth]{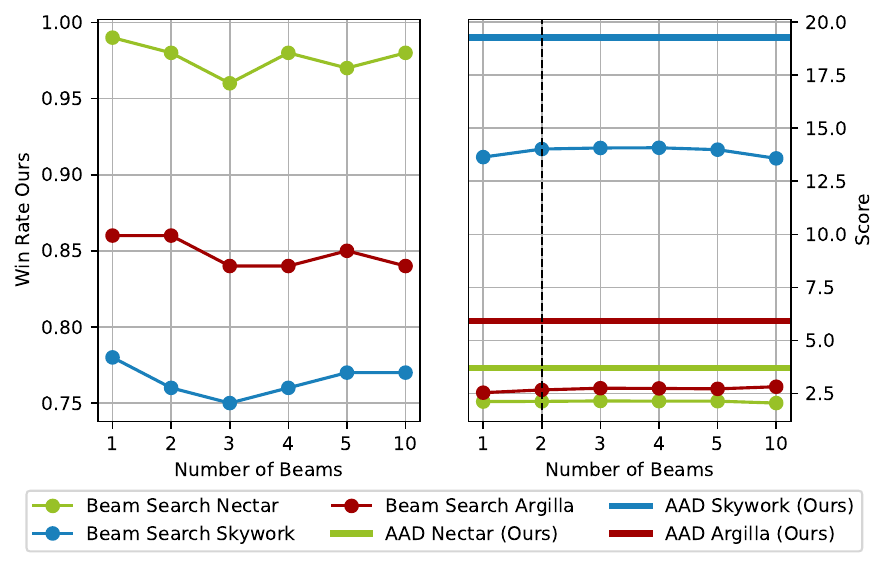}
     \caption{\textbf{Comparison of AAD against standard beam search}  using a 3B LLaMA model trained on the Argilla, Skywork, and Nectar datasets. The left panel reports the win rate of AAD against the corresponding dataset's beam search outputs across varying numbers of beams, showing that AAD consistently outperforms beam search regardless of beam width. The right panel presents the absolute scores, with AAD serving as a baseline for each dataset. While scores obtained with standard beam search vary slightly with the number of beams, increasing the beam width does not yield meaningful improvements in performance. In contrast, AAD produces stable and substantially higher scores across all settings, demonstrating its robustness and superior effectiveness compared to standard beam search.}
    \label{fig:beams_standard}
\end{figure}

\paragraph{Comparison with beam search.} In \cref{fig:beams_standard}, we compare standard beam search with AAD on the Argilla, Skywork, and Nectar models. The win rate curves show that AAD consistently surpasses beam search for all beam widths. The absolute scores further confirm that increasing the number of beams provides little to no benefit, while AAD achieves stable and clearly higher performance across all settings

\begin{figure}[ht]
        \centering
        \includegraphics[width=0.4\linewidth]{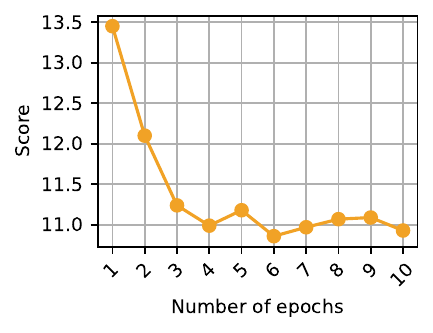}
        \caption{\textbf{Comparison of AAD across different stages of DPO training}, using a 3B LLaMA model trained on the Skywork dataset with LoRA finetuning and $\alpha = 0.1$. Performance decreases sharply after the first epoch but subsequently stabilizes, suggesting diminishing returns from additional training while maintaining overall consistency in the obtained scores.}
        \label{fig:epochs}
\end{figure}

\paragraph{Effect of overtraining.} In \cref{fig:epochs}, we study how the number of training epochs affects AAD. Performance drops after the first epoch but then remains stable, indicating that additional epochs offer limited gains while preserving consistent scores overall.

\paragraph{Data-scaling under full fine-tuning.} To verify that AAD's data-scaling behavior is not specific to LoRA, we replicate the experiment from \cref{perc} under full-parameter fine-tuning. We train a series of aligned models on increasing fractions of the Skywork preference data with a 3B Llama SFT base, and evaluate AAD against Bo2 on 800 held-out prompts. As shown in \cref{tab:scaling_ft}, the AAD win rate climbs from 51.1\% at 1\% of the data to 74.6\% at 100\%, and the AAD oracle reward exceeds the Bo2 oracle reward by at least 3.06 points from 5\% of the data onward. The pattern matches the LoRA setting in \cref{perc}, confirming that AAD's advantage under data scarcity is not an artifact of low-rank adaptation.

\begin{table}[h]
\centering
\small
\caption{\textbf{AAD versus Bo2 under full fine-tuning, across training-data fractions.} 3B LLaMA SFT trained on a fraction of the Skywork preference data and evaluated on 800 held-out prompts. $R(\textsc{aad})$ and $R(\textsc{bo2})$ are the oracle reward scores; the rightmost column is their difference.}
\label{tab:scaling_ft}
\begin{tabular}{ccccc}
\toprule
\textbf{Data \%} & \textbf{Win rate} & $R(\textsc{aad})$ & $R(\textsc{bo2})$ & \textbf{Score diff.} \\
\midrule
1   & 0.511 & $-2.59$ & $-2.11$ & $-0.48$ \\
5   & 0.599 &  $1.25$ & $-1.81$ &  $3.06$ \\
25  & 0.724 &  $6.11$ & $-0.89$ &  $7.00$ \\
50  & 0.726 &  $7.01$ & $-0.35$ &  $7.36$ \\
100 & 0.746 &  $8.36$ & $-0.56$ &  $8.92$ \\
\bottomrule
\end{tabular}
\end{table}

\paragraph{Latency and throughput.} We benchmark AAD across three single-GPU configurations and a two-GPU configuration relative to single-model greedy decoding. On a single GPU, AAD reaches roughly $0.5\times$ the throughput of greedy decoding on NVIDIA 3090, A6000, and A100, identical to EFT, which also performs two forward passes per token. As a concrete reference point, single-GPU AAD on OLMo-2-7B reaches $0.59\times$ the throughput of greedy decoding. With two GPUs and parallelized forward passes, AAD recovers to roughly $0.75\times$ throughput. Bo2 has higher aggregate token throughput but generates two full sequences in parallel, so the per-sequence cost is comparable to AAD. KV cache sharing and fused kernels are expected to reduce the overhead further.

\begin{table*}[ht]
\centering
\small
\setlength{\tabcolsep}{4pt}
\renewcommand{\arraystretch}{0.75}
\caption{\textbf{Performance of AAD on  fully finetuned DPO models}. Each cell shows reward ($R$) and win rate ($W$) of AAD against the corresponding method. Aligned models in this are trained with full finetuning instead of using a LoRA adapter like in the main results shown in \cref{tab:results_aad_lora}.}
\label{tab:results_aad_full}
\begin{tabularx}{\textwidth}{p{2.0cm}*{2}{>{\centering\arraybackslash}X>{\centering\arraybackslash}X}}
\toprule
\textbf{Method} & \multicolumn{4}{c}{\textbf{Models} \& \emph{Datasets}} \\
\cmidrule(lr){2-5}
& \multicolumn{2}{c}{\textbf{Llama 1B}}
& \multicolumn{2}{c}{\textbf{Llama 3B}} \\
 & $R$ & $W$ & $R$ & $W$ \\
\midrule\midrule
& \multicolumn{4}{c}{{\emph{Ultrafeedback}}} \\
\cmidrule(lr){2-5}
Greedy SFT   & -0.39 &0.72 &0.58  &0.77  \\
Greedy DPO   & -0.03 &0.65 & 1.04 & 0.7 \\
Bo2          & 0.18 &0.61 & 1.22 & 0.65 \\
EFT          &  0.3& 0.56& 0.5 & 0.83 \\
AAD (ours)   & \textbf{0.51} & - & \textbf{1.59} & - \\
\midrule
& \multicolumn{4}{c}{{\emph{Argilla}}} \\
\cmidrule(lr){2-5}
Greedy SFT   & 0.02 &0.85 &1.59  & 0.91 \\
Greedy DPO   & 1.65 &0.75 &3.64  &0.79  \\
Bo2          & 2.17 &0.72 & 4.06 & 0.76 \\
EFT          & 2.82 &0.58 & 5.01 & 0.56 \\
AAD (ours)   & \textbf{3.39} & - & \textbf{5.25} & - \\
\midrule
& \multicolumn{4}{c}{{\emph{OpenRLHF Mixture}}} \\
\cmidrule(lr){2-5}
Greedy SFT   & 2.06 &0.72 & 3.59 & 0.82 \\
Greedy DPO   & 3.15 & 0.63& 4.91 & 0.73 \\
Bo2          &  4.07&0.51 & 5.88 & 0.57 \\
EFT          & 3.64 &0.57 & 5.24 & 0.7 \\
AAD (ours)   & \textbf{4.04} & - & \textbf{6.26} & - \\
\midrule
& \multicolumn{4}{c}{{\emph{HHRLHF}}} \\
\cmidrule(lr){2-5}
Greedy SFT   & -1.91 &0.76 & -1.89 & 0.76 \\
Greedy DPO   &-0.63  & 0.64& 0.18 & 0.54 \\
Bo2          &  -0.75& 0.71& 0.09 & 0.57 \\
EFT          & \textbf{0.26} &0.3 & \textbf{0.47} & 0.35 \\
AAD (ours)   & -0.06 & - & 0.29 & - \\
\midrule
& \multicolumn{4}{c}{{\emph{Skywork}}} \\
\cmidrule(lr){2-5}
Greedy SFT   & -0.95 & 0.6 &7.93  &0.72  \\
Greedy DPO   & 1.12 &0.51 & 11.5 & 0.61 \\
Bo2          & 0.47 &0.57 & 11.71 &0.63  \\
EFT          & \textbf{2.00} & 0.48& 12.12 & 0.56 \\
AAD (ours)   & 1.55 & - & \textbf{13.4} & - \\
\midrule
& \multicolumn{4}{c}{{\emph{Nectar}}} \\
\cmidrule(lr){2-5}
Greedy SFT   & -0.26 &0.98 & 0.72 & 0.98 \\
Greedy DPO   & 1.32 &0.91 & 2.46 & 0.89 \\
Bo2          & 2.28 &0.77 & 2.9 & 0.79 \\
EFT          & 2.68 &0.65 & 3.35 & 0.58 \\
AAD (ours)   & \textbf{3.05} & - & \textbf{3.45} & - \\
\bottomrule
\end{tabularx}
\end{table*}

\begin{figure}[ht]
    \centering
    \includegraphics[width=0.7\linewidth]{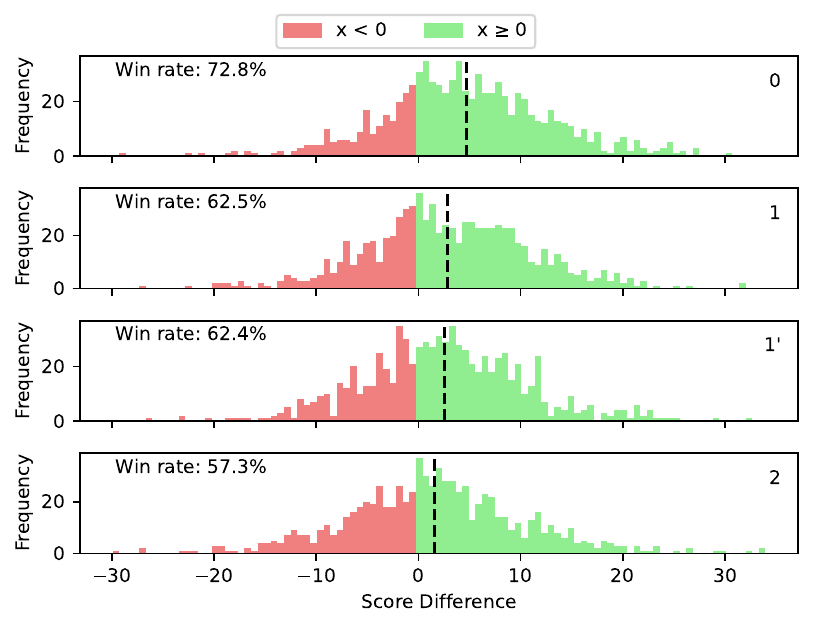}
    \caption{Histograms of score differences between AAD and Bo2 sampling across successive iterations of iterative DPO. The original model shows a clear advantage for AAD, with a win rate of 72.8\%. As iterations progress, the distributions gradually converge, and the win rate of AAD decreases. This occurs because AAD provides a more stable decoding method, while Bo2 sampling benefits substantially from iterative DPO retraining. As a result, the gap between the two methods narrows with additional iterations.}
    \label{histogram}
\end{figure}

\paragraph{Iterative DPO}
In this section, we highlight an additional property of Iterative DPO discussed in \cref{iter}. \Cref{histogram} presents histograms for the individual iterations, illustrating the score differences between AAD and Bo2 sampling.

\paragraph{Iterative DPO on additional preference datasets.} To assess whether the iterative DPO gains in \cref{iterative} generalize across datasets, we replicate the protocol on UltraFeedback (10\% subsample) and Argilla (5\% subsample). \cref{tab:iterative_extra} reports win rates against Iter 0 (the original DPO model trained on the subsample) with the corresponding $R$ scores in parentheses. Iter 1' (training from the original SFT checkpoint on AAD-generated data) produces the strongest models on both datasets, with AAD reaching an $R$ score of 2.21 on UltraFeedback and 2.25 on Argilla. Iter 2 degrades for all three decoding strategies, indicating that a single round of AAD-driven data generation is the practical sweet spot, consistent with known saturation effects in iterative self-improvement.

\begin{table*}[h]
\centering
\small
\setlength{\tabcolsep}{4pt}
\renewcommand{\arraystretch}{0.85}
\caption{\textbf{Iterative DPO on additional preference datasets.} Win rate against Iter 0 in percent, with $R$ score in parentheses, for Greedy, Bo2, and AAD decoding across three iterations. UltraFeedback uses a 10\% subsample of the preference data. Argilla uses 5\%.}
\label{tab:iterative_extra}
\begin{tabularx}{\textwidth}{p{1.5cm}*{2}{>{\centering\arraybackslash}X>{\centering\arraybackslash}X>{\centering\arraybackslash}X}}
\toprule
\textbf{Iteration} & \multicolumn{3}{c}{\emph{UltraFeedback (10\%)}} & \multicolumn{3}{c}{\emph{Argilla (5\%)}} \\
\cmidrule(lr){2-4} \cmidrule(lr){5-7}
& Greedy & Bo2 & AAD & Greedy & Bo2 & AAD \\
\midrule
Iter 1  & 57.6 ($-0.42$) & 57.9 ($0.19$) & 61.8 ($1.33$) & 52.1 ($-0.12$) & 54.0 ($0.69$) & 32.8 ($0.47$) \\
Iter 1' & 62.0 ($-0.20$) & 60.1 ($0.41$) & \textbf{72.9 ($2.21$)} & 62.9 ($0.44$) & 66.5 ($1.39$) & \textbf{62.9 ($2.25$)} \\
Iter 2  & 49.9 ($-0.73$) & 47.4 ($-0.27$) & 36.7 ($-0.27$) & 38.4 ($-0.62$) & 43.0 ($0.25$) & 14.9 ($-2.13$) \\
\bottomrule
\end{tabularx}
\end{table*}

\subsection{Code and models}
\label{github}
For reproducibility, the source code associated with this study can be accessed at:
\begin{center}
\href{https://github.com/ETH-DISCO/alignment-aware-decoding}{\texttt{https://github.com/ETH-DISCO/alignment-aware-decoding}}
\end{center}

\subsection{Additional Qualitative Examples}
\begin{figure}[ht]
    \centering
    \includegraphics[trim=0 480 0 0, clip]{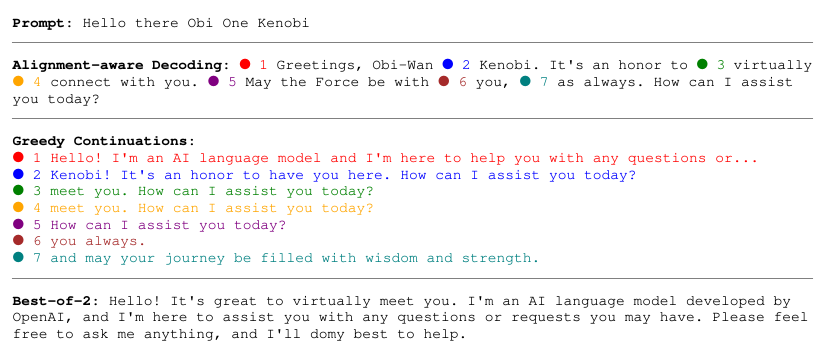}
     \caption{\textbf{Qualitative comparison of decoding strategies}. AAD recognizes the situational context and produces an appropriate, themed reply that includes the iconic phrase “May the Force be with you.” Greedy continuations are generated by iteratively feeding the prompt together with the current AAD prefix back into the model and greedily selecting the next token, revealing where the greedy trajectory diverges from AAD. These continuations tend to be generic or fragmentary and, from the raw prompt alone, do not seem to understand the context. Bo2 also produces a generic, non-contextual reply. It may appear slightly more coherent than a single greedy continuation but still fails to capture the situational or thematic cues present in the prompt. This highlights AAD’s strength in generating contextually aligned, thematically appropriate continuations.}
    \label{obiwan}
\end{figure}

\begin{figure}[ht]
    \centering
    \includegraphics[trim=0 395 0 0, clip]{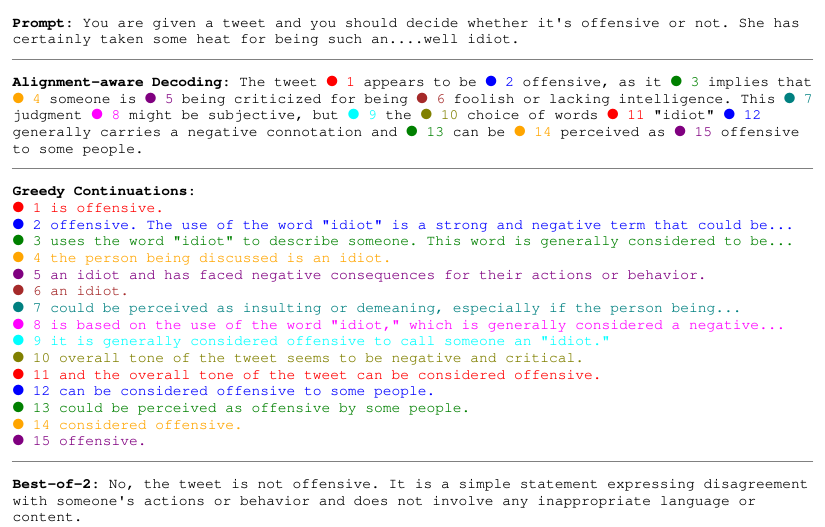}
     \caption{\textbf{Qualitative comparison of decoding strategies}. Greedy continuations are generated by iteratively feeding the prompt together with the current AAD prefix back into the model and greedily selecting the next token, revealing where the greedy trajectory diverges from AAD. AAD concludes that the tweet can be perceived as offensive, grounding this in the negative connotation of the word “idiot” and acknowledging that offensiveness is partly subjective. Greedy Continuations, when provided solely with the prompt, do not yield any explanation at all and are therefore not helpful for this task. Bo2, in contrast, judges the tweet as not offensive, treating it as simple disagreement rather than insult. }
    \label{tweet}
\end{figure}

\begin{figure}[ht]
    \centering
    \includegraphics[trim=0 450 0 0, clip]{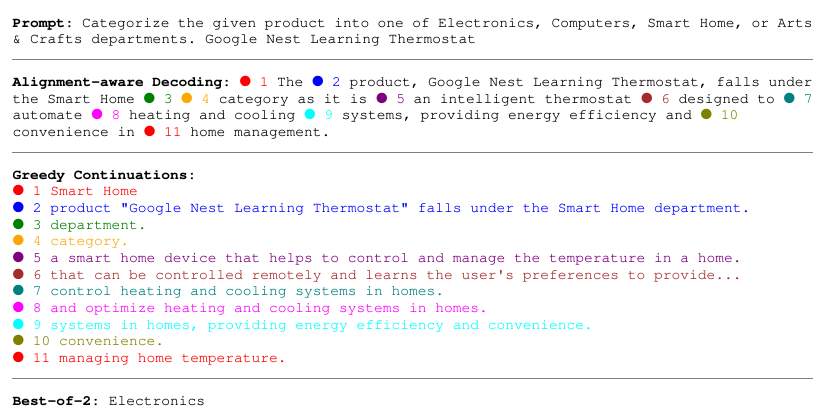}
    \caption{\textbf{Qualitative comparison of decoding strategies}. AAD produces an interpretable explanation, correctly assigning the product to Smart Home based on its function as an intelligent thermostat that automates heating and cooling systems. Greedy continuations are generated by iteratively feeding the prompt together with the current AAD prefix back into the model and greedily selecting the next token, revealing where the greedy trajectory diverges from AAD. Unlike AAD, greedy decoding does not provide coherent justifications. When applied to the raw prompt alone, it yields only short category labels without explanatory reasoning. Bo2 decoding misclassifies the product as Electronics and does not give any explanation. This comparison highlights the advantage of AAD in helpfulness.}
    \label{google_nest}
\end{figure}

\begin{figure}[ht]
    \centering
    \includegraphics[trim=0 290 0 0, clip]{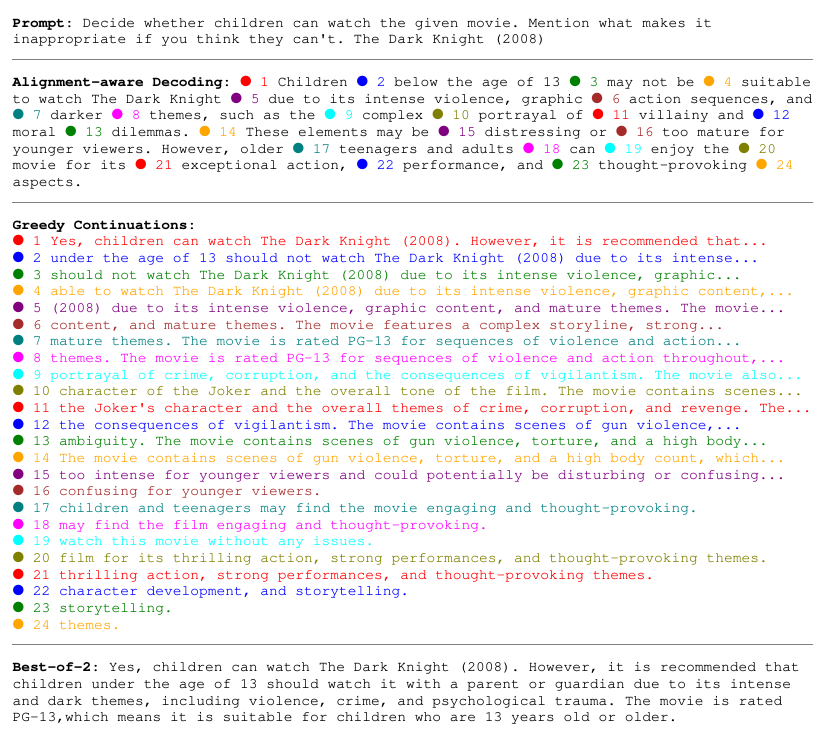}
     \caption{\textbf{Qualitative comparison of decoding strategies}. Greedy continuations are generated by iteratively feeding the prompt together with the current AAD prefix back into the model and greedily selecting the next token, revealing where the greedy trajectory diverges from AAD. AAD provides a balanced and contextually grounded judgment, explicitly noting that children under 13 may not be suitable viewers due to intense violence, graphic action sequences, darker themes, and complex moral dilemmas. Importantly, AAD also contrasts this with how older teenagers and adults may appreciate the film for its action, performances, and thought-provoking elements. Greedy continuations, when provided only with the prompt, lead to the misleading conclusion that children can watch the movie. Bo2 yields a generally correct but shallow response. It captures the broad conclusion but lacks the detailed reasoning, moral framing, and contextual awareness that make AAD’s output genuinely informative and situationally aligned. }
    \label{dk}
\end{figure}

In \cref{google_nest,dk,obiwan,tweet} we provide additional qualitative examples showing the benefits of AAD decoding.


\begin{table}[t]
\centering
\small
\caption{
We investigate the use of Proximal Policy Optimization (PPO) in combination with our AAD decoding strategy. For reward feedback during training, we employ the corresponding Picker Reward Models, each trained on 10\% of the corresponding dataset. In all experiments, the underlying model is a Llama SFT model~\citep{llama32}, and PPO training is conducted for 20,000 episodes. Our results indicate that the effectiveness of this approach varies across datasets: while PPO with AAD yields improvements on Nectar and Argilla, it underperforms on Skywork, highlighting the method’s inconsistency.
}

\label{tab:ppo}
\begin{tabularx}{\linewidth}{l *{6}{>{\centering\arraybackslash}X}}
\toprule
\textbf{Method} & \multicolumn{2}{c}{\emph{Argilla}} & \multicolumn{2}{c}{\emph{Skywork}} & \multicolumn{2}{c}{\emph{Nectar}} \\
\cmidrule(lr){2-3} \cmidrule(lr){4-5} \cmidrule(lr){6-7}
 & $R$ & $W$ & $R$ & $W$ & $R$ & $W$ \\
\midrule
GREEDY DPO & 2.09  & 0.78 & 7.22 & 0.29 & 1.79 & 0.8 \\
Bo2 & 2.51 & 0.78 & 10.4  & 0.16 & 2.35 & 0.64 \\
EFT & 3.33 & 0.7 & 3.48 & 0.47 & 2.49 & 0.57 \\
AAD (ours) & \textbf{4.46} & - & \textbf{3.02} & - & \textbf{2.68} & - \\
\bottomrule
\end{tabularx}
\end{table}

\end{document}